\documentclass[10pt,conference]{IEEEtran}

\newif\ifieee
\ieeefalse

\newif\ifarxiv
\arxivtrue

\ifieee
    \ifarxiv
        \PackageError{Configuration}
        {IEEE and arXiv cannot both be enabled}
        {Set either \string\ieeefalse\space or \string\arxivfalse.}
    \fi
\fi

\usepackage{multirow} %
\usepackage{makecell}
\usepackage{arydshln}
\usepackage[normalem]{ulem}
\usepackage{balance}
\usepackage{pgfplots}
\usepackage{newtxtext} %

\usepackage[sort,compress]{cite}

\usepackage{graphicx}

\usepackage[cmex10]{amsmath}
\usepackage{textcomp} %
\usepackage{gensymb} %
\usepackage{amsthm}

\usepackage{algorithmic}

\usepackage{array}

\ifCLASSOPTIONcompsoc
  \usepackage[caption=false,font=normalsize,labelfont=sf,textfont=sf,farskip=0pt]{subfig}
\else
  \usepackage[caption=false,font=footnotesize,farskip=0pt]{subfig}
\fi

\usepackage[acronym]{glossaries}
\usepackage{xspace}
\usepackage[hyphens]{url}
\usepackage{booktabs}
\usepackage[dvipsnames]{xcolor}%
\usepackage[utf8]{inputenc}
\usepackage[T1]{fontenc}
\usepackage{multirow}
\usepackage{diagbox}
\newif\ifieee
 \ieeefalse

\ifieee
\else
\usepackage[colorlinks=true,allcolors=black]{hyperref} %
\fi

\usepackage[capitalise]{cleveref} %

\newacronymstyle{long-short-br}
{%
  \GlsUseAcrEntryDispStyle{long-short}%
}%
{%
  \GlsUseAcrStyleDefs{long-short}%
}
\setacronymstyle{long-short-br}

\usepackage{transparent}
\usepackage{tikz}
\ifarxiv
    \newcommand\copyrighttext{%
      \scriptsize Accepted for presentation at SIBGRAPI 2026. The final published version will be available on IEEE~Xplore.}
    \newcommand\copyrightnotice{%
    \begin{tikzpicture}[remember picture,overlay]
    \node[anchor=south,yshift=30pt,xshift=0pt] at (current page.south) {\fbox{\transparent{0.85}\parbox{\dimexpr0.6\textwidth-\fboxsep-\fboxrule\relax}{\copyrighttext}}};
    \end{tikzpicture}%
    }
\else
\fi

\usepackage[table]{xcolor}
\usepackage{booktabs}
\usepackage{array}

\definecolor{StrongRed}{HTML}{B2182B}
\definecolor{LightRed}{HTML}{F4A3A8}
\definecolor{White}{HTML}{FFFFFF}
\definecolor{LightGreen}{HTML}{B8E186}
\definecolor{StrongGreen}{HTML}{4D9221}

\newif\iffinal
\finaltrue
\newcommand{\cmtid}{229}

\iffinal
\else
\usepackage[switch]{lineno}
\fi

\newcommand*{\RL}[2][]{\textcolor{Rhodamine}{[\textbf{\ifthenelse{\equal{#1}{}}{RL}{RL(#1)}}: #2]}}
\newcommand*{\DM}[2][]{\textcolor{orange}{[\textbf{\ifthenelse{\equal{#1}{}}{DM}{DM(#1)}}: #2]}}
\newcommand*{\GL}[2][]{\textcolor{brown}{[\textbf{\ifthenelse{\equal{#1}{}}{GL}{GL(#1)}}: #2]}}
\newcommand*{\AD}[2][]{\textcolor{cyan}{[\textbf{\ifthenelse{\equal{#1}{}}{AD}{AD(#1)}}: #2]}}

\newcommand\major[1]{{#1}}

\newcommand\blue[1]{{\textcolor{black}{#1}}}

\newcounter{fncounter}
\begin{document}

\iffinal
    \newcommand{\urlSupplementary}{\url{https://github.com/UFPR-IPASP-PR/3D-Vision-Benchmark/}}
\fi

\title{Evaluating 2D and 3D-Aware Vision Foundation Models for Vehicle Attribute Recognition}

\iffinal

\author{
\IEEEauthorblockN{Alexandre V. Delazeri\IEEEauthorrefmark{1}, Gabriel E. Lima\IEEEauthorrefmark{1}, Eduil Nascimento~Jr.\IEEEauthorrefmark{2}, Rayson~Laroca\IEEEauthorrefmark{3}$^,$\IEEEauthorrefmark{1}, and David Menotti\IEEEauthorrefmark{1}}
\IEEEauthorblockA{
    \IEEEauthorrefmark{1}Department of Informatics, Federal University of Paran\'{a}, Curitiba, Brazil \\
    \IEEEauthorrefmark{2}Department of Technological Development and Quality, Paran\'{a} Military Police, Curitiba, Brazil \\
    \IEEEauthorrefmark{3}Graduate
    Program in Informatics, Pontifical Catholic University of Paran\'a, Curitiba, Brazil \\
        \hspace{-0.75mm}\IEEEauthorrefmark{1}\hspace{-0.35mm}\texttt{\small{\{avdelazeri,gelima,menotti\}}@inf.ufpr.br\quad \IEEEauthorrefmark{2}\hspace{0.1mm}{\tt\small {eduiljunior}@pm.pr.gov.br} \quad \IEEEauthorrefmark{3}\hspace{0.1mm}{\tt\small rayson@ppgia.pucpr.br}}
    }
}

\fi

\maketitle

\ifarxiv
    \copyrightnotice
\else
\fi

\newacronym{alpr}{ALPR}{Automatic License Plate Recognition}
\newacronym{capes}{CAPES}{Coordination for the Improvement of Higher Education Personnel}
\newacronym{cls}{CLS}{Class Token}
\newacronym{cnn}{CNN}{Convolutional Neural Network}
\newacronym{cnpq}{CNPq}{National Council for Scientific and Technological Development}
\newacronym{croco}{CroCo}{Cross-view Completion}
\newacronym{fgvc}{FGVC}{Fine-Grained Vehicle Classification}
\newacronym{gap}{GAP}{Globally Average-pooled Patch}
\newacronym{ma-acc}{Ma-Acc}{Macro-Accuracy}
\newacronym{mi-acc}{Mi-Acc}{Micro-Accuracy}
\newacronym{ood}{OOD}{Out-of-Distribution}
\newacronym{senatran}{SENATRAN}{Brazil's National Traffic Secretariat}
\newacronym{vit}{ViT}{Vision Transformer}

\ifieee
\vspace{-3.575mm}
\else
\vspace{-3.575mm}
\fi
\begin{abstract}

Vehicle attribute recognition is an important task in intelligent transportation systems, particularly when Automatic License Plate Recognition~(ALPR) is unavailable or unreliable. Although vision foundation models have shown strong transferability across domains, their effectiveness for fine-grained vehicle classification remains underexplored. Moreover, given the inherently three-dimensional structure of vehicles, it is unclear whether emerging 3D-aware foundation models offer advantages over standard 2D architectures. This paper presents an empirical benchmark of \major{14} state-of-the-art 2D and 3D-aware vision foundation models. Using the challenging real-world UFPR-VeSV dataset, we evaluate these models as frozen feature extractors via linear probing for vehicle type, make, and model recognition. We further stress-test the best-performing models under few-shot learning and Out-of-Distribution~(OOD) domain shifts. Our results show that standard 2D self-supervised models, particularly DINOv3, substantially outperform 3D-aware models in fine-grained tasks, achieving over $93$\% Macro-Accuracy for make and model recognition. \major{However, the 3D-aware Depth Anything~v2 exhibits stronger invariance to viewing angles in vehicle type classification}. These findings motivate hybrid approaches that combine 2D and 3D priors for robust vehicle recognition.
Our code is publicly available at \textit{\urlSupplementary}.

\end{abstract}

\IEEEpeerreviewmaketitle

\section{Introduction}
\label{sec:introduction}

\glsresetall

Vehicle recognition is a core component of intelligent transportation systems, supporting applications such as traffic monitoring, parking management, and forensic analysis~\cite{laroca2025advancing,he2024vehicle}.
Although retrieval frameworks have relied on \gls{alpr}, the performance of this approach degrades in real-world surveillance settings.
Factors such as partial occlusion, viewpoint variation, and poor image quality can render license plates illegible~\cite{wojcik2025lplc,nascimento2025toward,laroca2026competition}.

To address this limitation, vehicle attribute recognition can serve as a complementary approach.
Identifying attributes such as vehicle type, make, and model from the vehicle's global appearance provides greater robustness to the adverse capture conditions described above~\cite{li2025weakly,lu2025automated,tan2025cross,orru2026revisiting}.
Unlike license plates, a vehicle's overall appearance and structure are not confined to a single localized~region.

Nevertheless, reliable vehicle attribute recognition in real-world surveillance remains challenging due to low interclass variance among similar models and high intraclass variance from changing viewpoints and illumination~\cite{lu2025automated,li2025weakly,tan2025cross}. 
Furthermore, these scenarios suffer from severe class imbalance, as a few popular vehicles dominate traffic~\cite{orru2026revisiting,senatran_frota_2026}.
Overcoming these obstacles demands highly discriminative and generalizable feature representations. 

Recently, vision foundation models have gained prominence by learning generalizable representations from massive datasets without task-specific training~\cite{simeoni2025dinov3}. 
While standard 2D architectures are highly effective for broad classification tasks, an emerging class of 3D-aware models explicitly embeds spatial and geometric priors to master multi-view consistency~\cite{dust3r_cvpr24}.
However, both paradigms remain underexplored for vehicle attribute recognition. 
Given the inherently three-dimensional structure of vehicles, it remains an open question whether these 3D-aware representations provide an advantage over 2D semantic priors in this domain.

To bridge this gap, we benchmark $5$ standard 2D and $9$ 3D-aware vision foundation models as frozen feature extractors via linear probing. 
Using a public surveillance dataset\cite{lima2026toward}, we assess their performance on vehicle type~(e.g., car, motorcycle), make~(e.g., Audi, Toyota), and model~(e.g., Audi A3, Toyota Corolla) recognition. 
\blue{Furthermore, we test the top-performing models via few-shot learning, \gls{ood} generalization, \major{and non-linear probing}.
Through these analyses, we aim to guide future model selection and architectural design for vehicle attribute recognition.}

The remainder of this paper is organized as follows. 
\cref{sec:related_work} reviews related work. 
\cref{sec:background} provides the background on 2D and 3D-aware vision foundation models. 
\cref{sec:methodology} details the experimental methodology, followed by the discussion of results in \cref{sec:results}. 
Finally, \cref{sec:conclusion} summarizes the key findings and outlines directions for future~work.

\section{Related Work}
\label{sec:related_work}

\blue{Early vehicle attribute recognition relied on handcrafted features and conventional classifiers~\cite{zhang2013reliable,ma2005edge}, sometimes leveraging 3D geometry to improve accuracy~\cite{chen2011vehicle}. 
The release of large-scale datasets, such as Stanford Cars~\cite{krause2013collecting}, CompCars~\cite{yang2015compcars}, and BIT-Vehicle~\cite{dong2015vehicle}, accelerated the shift toward deep learning. 
Modern methods primarily adopt \glspl{cnn} and \glspl{vit}, enhancing these architectures with part-based attention, multi-scale feature extraction, and metric learning~\cite{li2025weakly,tan2025cross,gayen2025simsanet}.}

Concurrently, several studies sought to incorporate explicit 3D vehicle information into deep learning frameworks.
These approaches relied on 3D bounding boxes~\cite{sochor2019boxcars,krause2013object} or aligned 3D object models to learn viewpoint-invariant features~\cite{lin2014jointly}.
\major{However, since obtaining 3D annotations is impractical for many real-world applications~\cite{yao2025labelany3d}, such approaches have gradually received less attention from the research~community\cite{GAYEN2024101885}}.

More recently, the community has started to explore vision foundation models for vehicle attribute recognition, inspired by their strong transferability across broader image classification domains~\cite{oquab2023dinov2}.   
\major{However, current approaches are still limited~\cite{sathyam2025foundation}}.
Studies often need to fine-tune models on target datasets, missing the benefits of leveraging frozen representations~\cite{wu2026perception,munoz2025veri}.
Furthermore, existing evaluations are almost exclusively confined to standard 2D architectures. 
Consequently, the potential of 3D-aware foundation models remains unexplored in this context.

\blue{Unlike previous studies, this work introduces a comprehensive benchmark that directly compares state-of-the-art 2D vision foundation models against emerging 3D-aware alternatives. 
By evaluating these models as frozen feature extractors, we assess how their representations transfer to vehicle attribute recognition under real-world surveillance conditions.
This analysis allows us to determine whether explicit 3D priors provide a tangible advantage over standard 2D foundation~models.}

\section{Background}
\label{sec:background}

\blue{Foundation models have emerged as a key paradigm for learning transferable representations, as they are pre-trained on massive datasets --- typically via self-supervision --- and can be adapted to downstream tasks~\cite{bommasani2022opportunitiesrisksfoundationmodels}. 
For image classification, these models can be evaluated by freezing the backbone to serve as a feature extractor~\cite{grill2020byol,caron2021dino}. 
A single linear classification head is then trained on top, leveraging the learned representations without requiring full network fine-tuning.}

Standard vision foundation models are commonly trained with objectives such as masked image modeling (e.g., MAE, iBOT~\cite{zhou2021ibot}) and matching embeddings across different image views (e.g., DINOv1/v2/v3~\cite{caron2021dino,oquab2023dinov2,simeoni2025dinov3}). 
These techniques teach the network to reconstruct missing image patches, group visually similar features, or align augmented views of the same image~\cite{simeoni2025dinov3}. 
Because they are trained on large image collections without explicit geometric supervision, their representations primarily capture appearance-based cues\footnote{Nevertheless, these models can implicitly encode spatial relationships~\cite{amir2021deep}.}.

\begin{figure}[t]
    \centering
    
    \resizebox{0.925\linewidth}{!}{
        \begin{tabular}{cccc}
            \scriptsize Reference &
            \scriptsize Target (masked) &
            \scriptsize Reconstruction &
            \scriptsize Target (original)\\
            
            \includegraphics[height=12.5ex]{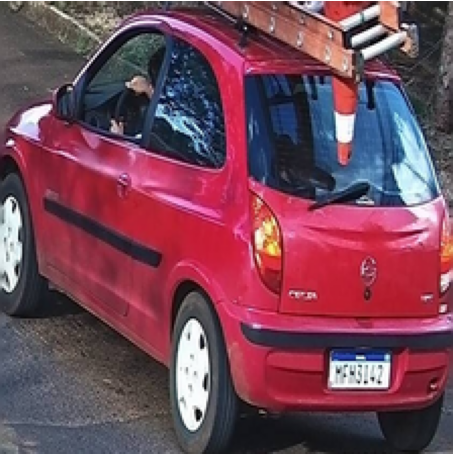} &
            \includegraphics[height=12.5ex]{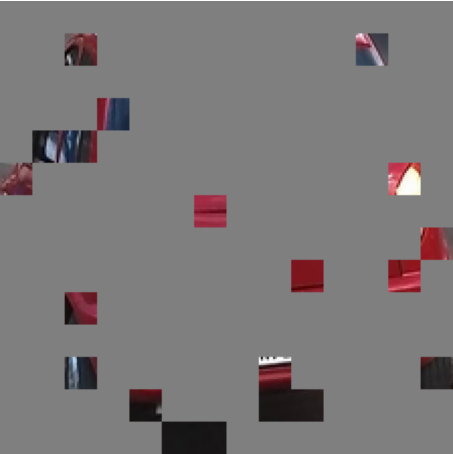} &
            \includegraphics[height=12.5ex]{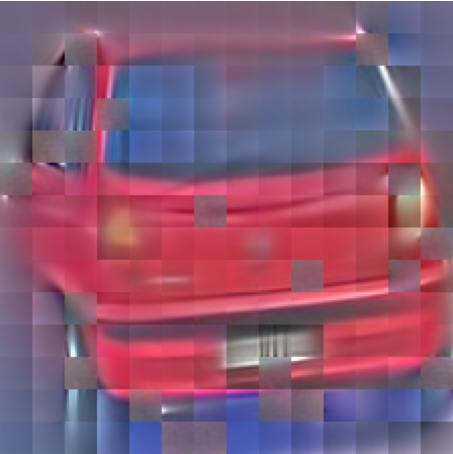} &
            \includegraphics[height=12.5ex]{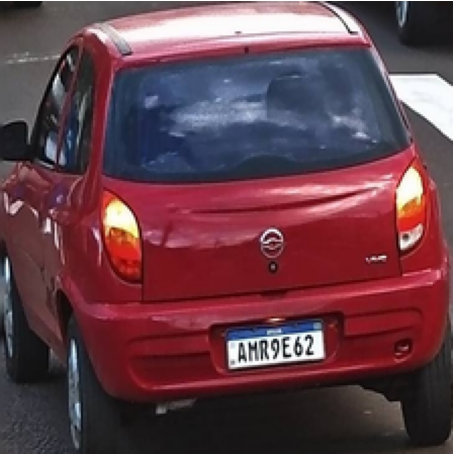} \\
            
            \includegraphics[height=12.5ex]{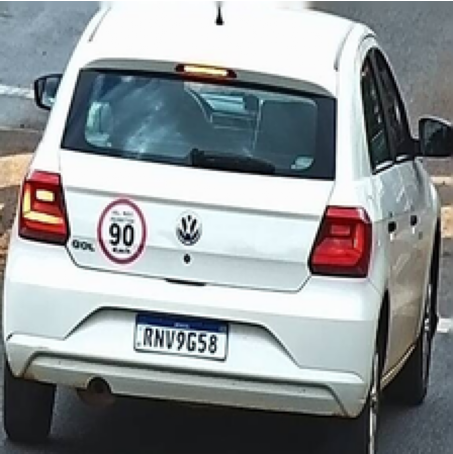} &
            \includegraphics[height=12.5ex]{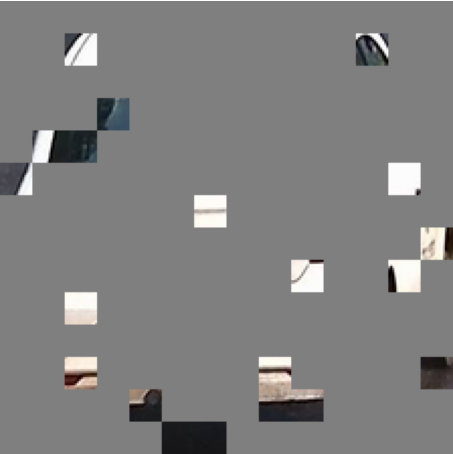} &
            \includegraphics[height=12.5ex]{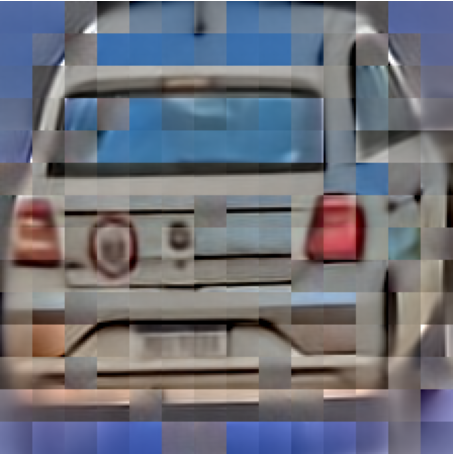} &
            \includegraphics[height=12.5ex]{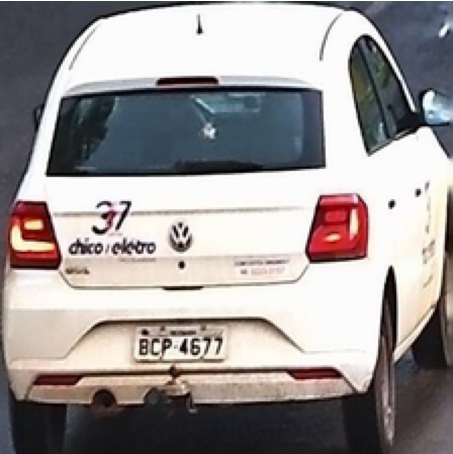} \\
        \end{tabular}
    }
    \vspace{-1mm}
    
    \caption{Illustration of the cross-view completion objective. \blue{The network receives a reference image alongside a masked target image from a different viewpoint. By mapping structural correspondences from the reference image to reconstruct the missing target patches, the network naturally learns viewpoint-invariant representations and 3D geometry}.}
    \label{fig:cross-view-completion-example}
\end{figure}

In contrast, 3D-aware vision foundation models incorporate self-supervised objectives that encourage geometric consistency, enabling the network to understand underlying object shapes. 
A strategy within this context is cross-view learning (e.g., CroCo v2~\cite{croco_v2}), which extends masked image modeling to paired images of the same scene captured from different viewpoints (see~\cref{fig:cross-view-completion-example}). 
Thus, the foundation model is compelled to learn viewpoint-invariant representations and infer occluded object structures.

Other related approaches focus on predicting 3D information from 2D images. 
For instance, depth-supervised strategies (e.g., Depth Anything models~\cite{depth_anything_v1,depth_anything_v2,depthanything3}) learn dense depth representations through large-scale pseudo-supervision, embedding a spatial hierarchy into the learned features. 
Similarly, multi-view reconstruction frameworks (e.g., DUSt3R~\cite{dust3r_cvpr24}, MASt3R~\cite{mast3r_eccv24}) learn geometry-aware representations by jointly estimating pixel correspondences, depth, and scene structure.

\blue{Finally, native 3D foundation models (e.g., Gamba~\cite{shen2025gamba}, Hunyuan3D~\cite{hunyuan3d2025hunyuan3d}, and SAM 3D~\cite{sam3dteam2025sam3d3dfyimages}) incorporate explicit 3D representations, including point clouds, meshes, gaussian splats, and voxels, directly into the learning process. 
By learning directly from geometric data, these models avoid image projections and better preserve the spatial scale and structural relationships of physical objects.
However, despite their greater geometric fidelity, model scalability remains limited by the scarcity of large-scale, diverse 3D~datasets.}

\section{Methodology}
\label{sec:methodology}

\blue{To assess the transferability of 2D and 3D-aware vision foundation models for vehicle attribute recognition, we benchmark \major{14} methods across vehicle type, make, and model recognition tasks via linear probing on frozen feature representations. 
Following this baseline, we subject the top-performing models from each paradigm to non-linear probing to further assess feature separability. 
Finally, we evaluate these models through few-shot data efficiency tests and \gls{ood} generalization to understand how distinct representational priors affect the model's representation robustness.}

\begin{figure}[!htb]
    \centering
    \resizebox{0.9\linewidth}{!}{
        \includegraphics[height=11.75ex]{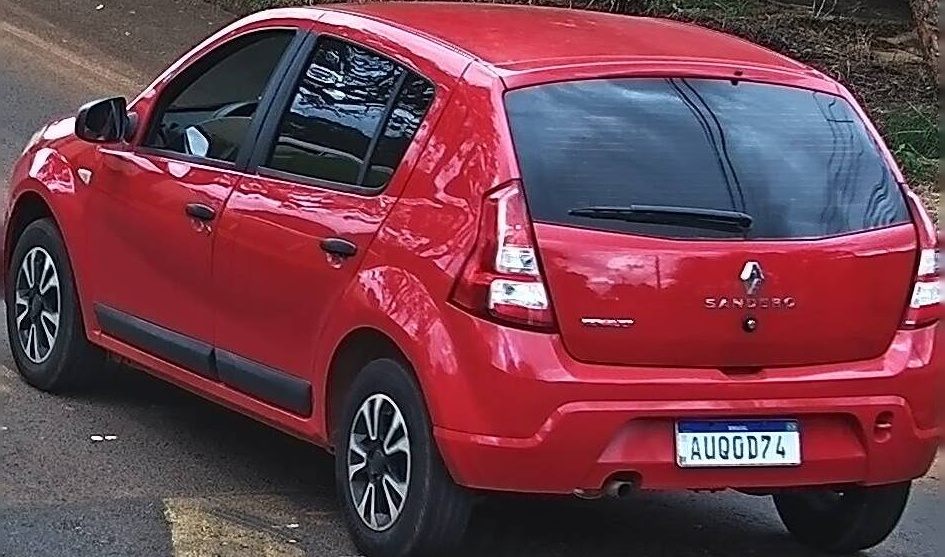}
        \includegraphics[height=11.75ex]{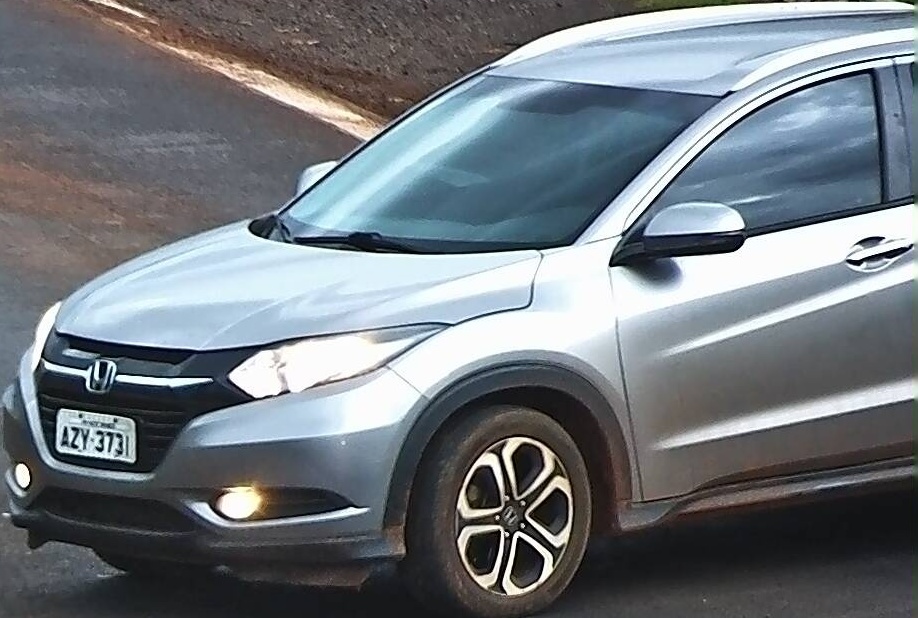}
        \includegraphics[height=11.75ex]{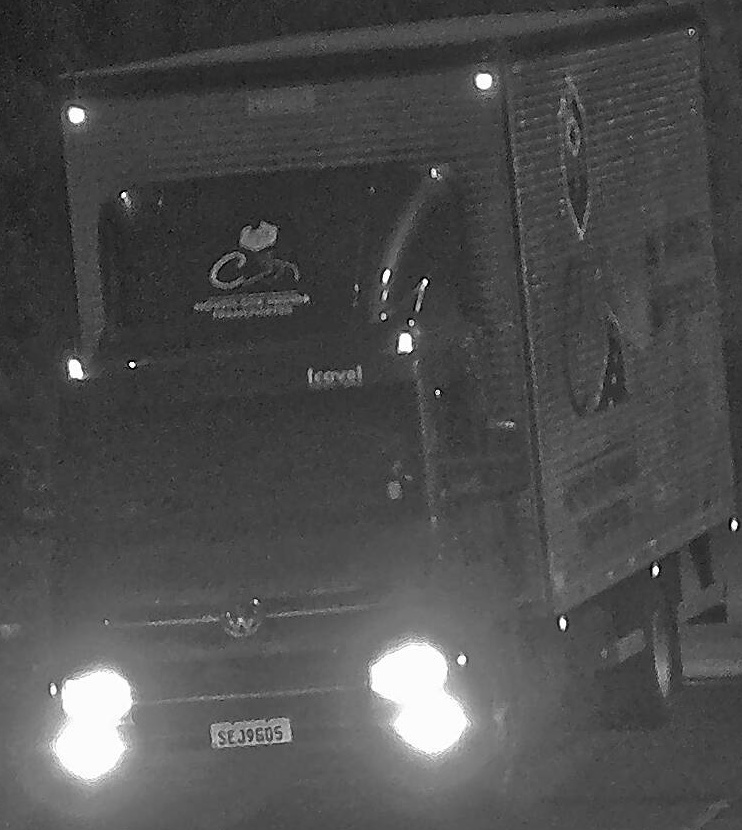}
    }%
    
    \vspace{1mm}
    
    \resizebox{0.925\linewidth}{!}{
        \includegraphics[height=11.75ex]{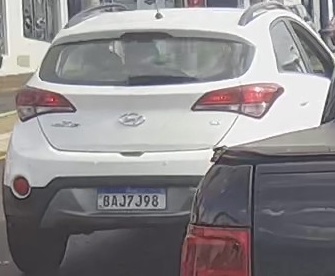}
        \includegraphics[height=11.75ex]{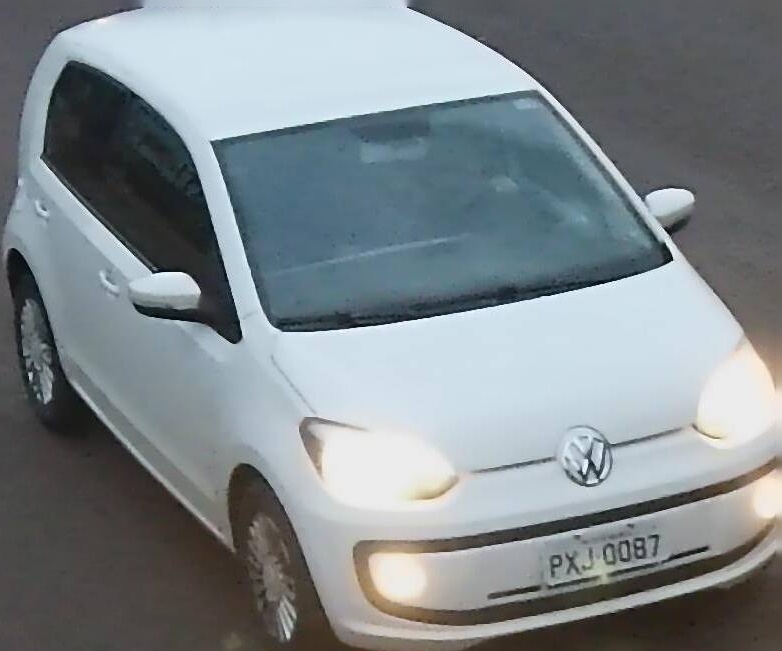}
        \includegraphics[height=11.75ex]{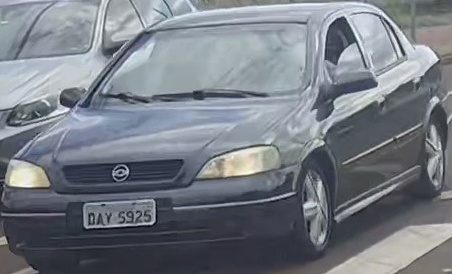}
    }%
    
    \vspace{1mm}
    
    \resizebox{0.925\linewidth}{!}{
        \includegraphics[height=11.75ex]{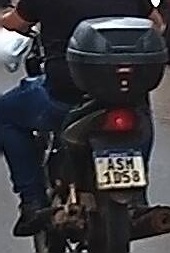}
        \includegraphics[height=11.75ex]{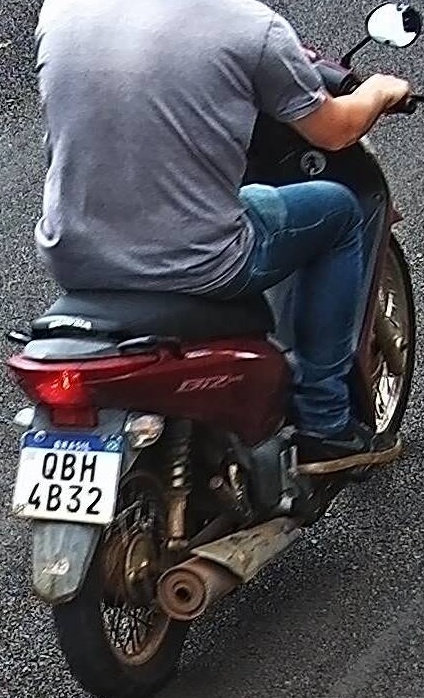}
        \includegraphics[height=11.75ex]{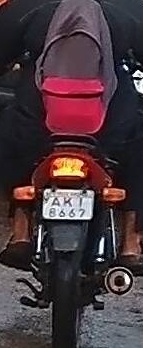}
        \includegraphics[height=11.75ex]{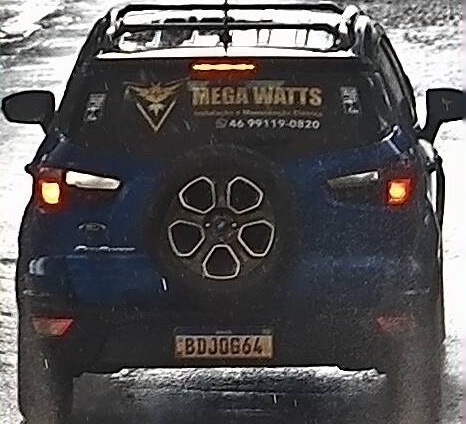}
        \includegraphics[height=11.75ex]{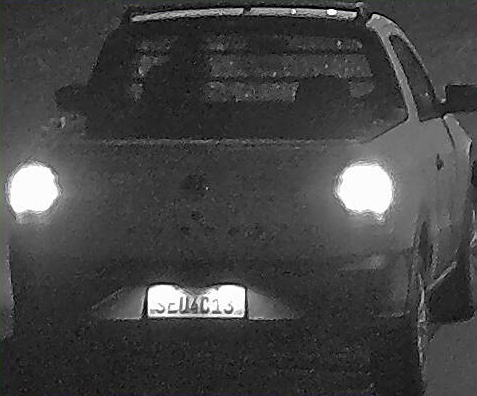}
    }
    
    \vspace{-2mm}
    
    \caption{Examples of adverse capturing surveillance conditions present in the UFPR-VeSV dataset~\cite{lima2026toward}. Image adapted from~\cite{lima2026toward}.}
    \label{fig:ufpr-vesv-dataset}
\end{figure}

\blue{We use the UFPR-VeSV dataset~\cite{lima2026toward}, a public collection of $24{,}945$ real-world traffic images from the Military Police of Paraná~(Brazil). 
It categorizes vehicles into $14$ types, $26$ makes, and $136$ models, presenting a severe long-tailed class distribution typical of real-world environments~\cite{orru2026revisiting,senatran_frota_2026}. 
We selected this dataset because: 
(i)~it provides ground-truth labels for all evaluated attributes in this work; 
(ii)~it captures adverse real-world surveillance conditions (e.g., partial occlusions, and illumination variations), as depicted in \cref{fig:ufpr-vesv-dataset}; and 
(iii)~its viewpoint annotations allow for orientation-based analyses.}

Following the UFPR-VeSV protocol~\cite{lima2026toward}, experiments are evaluated across the 10 official stratified train-val-test splits. 
We report the mean \gls{mi-acc} and \gls{ma-acc} across all folds. \major{Micro accuracy measures overall accuracy across all samples, whereas macro accuracy averages per-class accuracy, weighting all classes equally to prevent majority classes from masking poor generalization on minority classes under severe class imbalance.}

\blue{The following subsections detail our experimental methodology. 
\cref{subsec:benchmark} introduces the linear probing baseline and the selected foundation models. 
\cref{subsec:represetantion-analysis} outlines the robustness analyses of the model's learned~representations.} 
\major{To ensure reproducibility, the code, along with the specifications of checkpoints and weights used for evaluated models, is publicly available at \textit{\urlSupplementary}}.

\subsection{Selected Models and Benchmark Setup}
\label{subsec:benchmark}

\cref{tab:selected-vision-foundation-models} summarizes the evaluated models, chosen based on two key criteria. 
First, we prioritized \gls{vit}-Large backbones to isolate the impact of the pre-training objective from architectural capacity. 
Second, most 3D-aware models build upon DINOv2, allowing us to measure the direct impact of adding geometric priors to a 2D baseline.

\begin{table*}[!hbt]
    \centering
    \caption{Overview of the evaluated 2D and 3D-aware vision foundation models. The table details their respective backbones, pre-training tasks, and feature extraction strategies.}%
    \label{tab:selected-vision-foundation-models}

    \vspace{-2mm}
    
    \resizebox{0.9\linewidth}{!}{
        \begin{tabular}{@{} c c l l l r l r @{}}
                \toprule
                Paradigm & Year & Model & Backbone & Pre-training Objective & Params. & Latent Target & Dim. \\
                \midrule
                \multirow{6}{*}{Standard 2D} 
                & 2021 & DINOv1\cite{caron2021dino}                 & ViT-B/16          & Self-distillation                         & 85M           & [CLS] + GAP       & $1{,}536$     \\
                & 2022 & iBOT\cite{zhou2021ibot} (ImageNet-22K)     & ViT-L/16          & Masked Image Modeling                     & 307M          & [CLS] + GAP       & $2{,}048$     \\
                & 2023 & DINOv2\cite{oquab2023dinov2}               & ViT-L/14          & Self-distillation                         & 300M          & [CLS] + GAP       & $2{,}048$     \\
                & 2024 & DINOv3\cite{simeoni2025dinov3}             & ViT-L/16          & Self-distillation                         & 300M          & [CLS] + GAP       & $2{,}048$     \\
                & 2024 & SAM 3\cite{carion2025sam3segmentconcepts}  & Hiera-L/14        & Promptable Segmentation                   & $\sim$300M    & GAP               & $1{,}024$     \\
                \midrule
                \multirow{8}{*}{3D-Aware} 
                & 2023 & CroCo v2\cite{croco_v2}                    & ViT-L/14          & Cross-view Completion                     & 303M          & GAP               & $768$         \\
                & 2024 & Depth Anything v1\cite{depth_anything_v1}   & ViT-L/14          & Self-distillation + Monocular Depth       & 300M          & [CLS] + GAP       & $2{,}048$     \\
                & 2024 & Depth Anything v2\cite{depth_anything_v2}   & ViT-L/14          & Self-distillation + Monocular Depth       & 300M          & [CLS] + GAP       & $2{,}048$     \\
                & 2024 & DUSt3R (Encoder)\cite{dust3r_cvpr24}       & ViT-L/14          & Cross-view Completion + Point Cloud Reg.  & 303M          & GAP               & $768$         \\
                & 2024 & Gamba\cite{shen2025gamba}                  & DINOv2 + Mamba    & Explicit 3D Reconstruction                & 424M          & Final latent      & $16{,}384$    \\
                & 2024 & Hunyuan3D 2.1\cite{hunyuan3d2025hunyuan3d} & DINOv2 + DiT      & Explicit 3D Reconstruction                & 3250M         & Geometry tokens   & $4{,}096$     \\
                & 2024 & MASt3R\cite{mast3r_eccv24} (Encoder)       & ViT-L/14          & Cross-view Completion + Stereo Matching   & 303M          & GAP               & $768$         \\
                & 2024 & SAM 3D\cite{sam3dteam2025sam3d3dfyimages}  & DINOv2 + MoT/14   & Explicit 3D Reconstruction                & 3000M         & Shape latent      & $4{,}096$     \\
                & 2025 & Depth Anything 3\cite{depthanything3}      & ViT-L/14          & Self-distillation + Monocular Depth       & 300M          & CAM + GAP         & $2{,}048$     \\
                \bottomrule
            \end{tabular}
    }
\end{table*}

\blue{For a fair comparison, we evaluate all models as frozen feature extractors, as fine-tuning could mask underlying representational shortcomings. 
Specifically, we extract embeddings from the encoder's highest-level latent representation. 
For 3D reconstruction models, features are extracted before the spatial regression or generation modules to ensure they capture generalized visual and geometric understanding rather than rendering-specific details.}

\blue{For standard Vision Transformers, we concatenate the final \gls{cls} and \gls{gap} tokens. 
For models with specialized tokens (e.g., the CAM token in Depth Anything~3), this token replaces the CLS token, whereas for architectures lacking dedicated tokens (e.g., SAM~3, CroCo v2, DUSt3R), we apply global average pooling directly to the final latent sequence. 
These features are extracted offline to train independent linear classifiers for vehicle type, make, and model.}

\blue{All probes are trained directly on the native feature spaces without dimensionality reduction. 
We optimize these probes using Adam (batch size of~$32$, learning rate of $10^{-4}$, weight decay of $10^{-4}$) for up to $100$ epochs. 
We also employ early stopping with a $5$-epoch patience. 
Finally, to address class imbalance, all probes are trained using an inverse-frequency weighted cross-entropy loss.}

\major{Finally, we evaluate the best-performing 2D and 3D-aware models using a non-linear probing setup to determine whether 3D pretraining yields feature spaces that are informative yet not linearly separable.
Specifically, we replace the linear classifier with a multi-layer perceptron featuring a single $512$-neuron hidden layer, with $0.2$ dropout and keeping all other training hyperparameters identical to the linear baseline.}

\subsection{Representation Robustness Analysis}
\label{subsec:represetantion-analysis}

\blue{
Real-world traffic surveillance systems frequently encounter operational constraints, such as domain shifts and a scarcity of annotated data for rare vehicles. 
To evaluate feature robustness under these conditions, we assess the top-performing 2D and 3D-aware architectures on data efficiency and \gls{ood} generalization. 
We restrict these analyses to vehicle type recognition to maintain a unified scope.}

\major{To evaluate knowledge transfer under limited data, we simulate low-resource scenarios by retaining $1$\%, $5$\%, $10$\%, and $25$\% of the original training instances across the 10 official UFPR-VeSV splits. 
To prevent the omission of minority classes, we ensure that at least one sample per class is always retained. 
The linear probes are then trained from scratch on these subsets and evaluated on the original test sets.}

\major{To test robustness against domain shifts, we evaluate the models on a curated \gls{ood} set of $200$ unseen vehicle images (see samples in Fig. 3). 
Captured by a similar surveillance infrastructure but representing a different data distribution, this set approximates the UFPR-VeSV original type class distribution. 
Using the probes trained in \cref{subsec:benchmark}, we investigate whether 3D-aware representations are less prone to misclassification on unfamiliar vehicles than the 2D methods.}

\begin{figure}[!htb]
    \centering
    \resizebox{0.925\linewidth}{!}{
        \includegraphics[height=11.75ex]{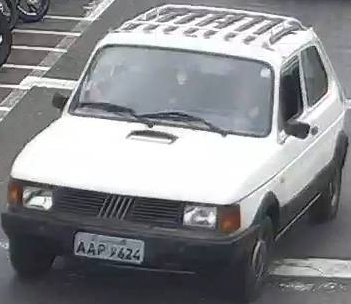}
        \includegraphics[height=11.75ex]{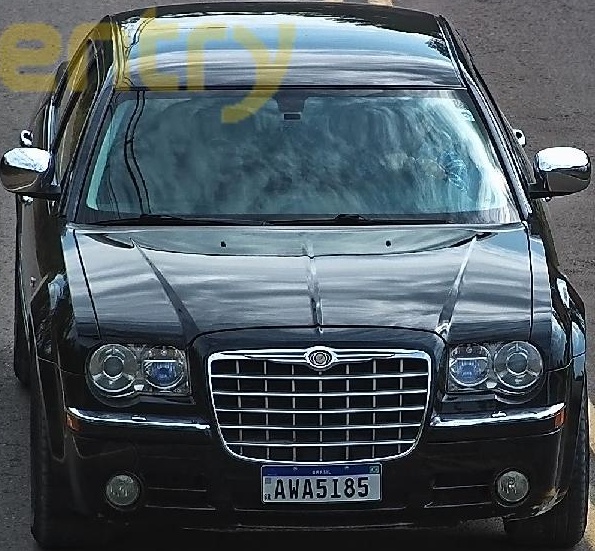}
        \includegraphics[height=11.75ex]{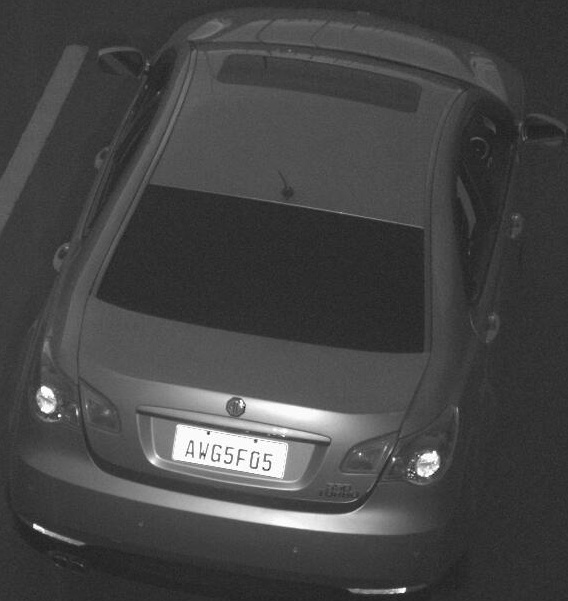}
        \includegraphics[height=11.75ex]{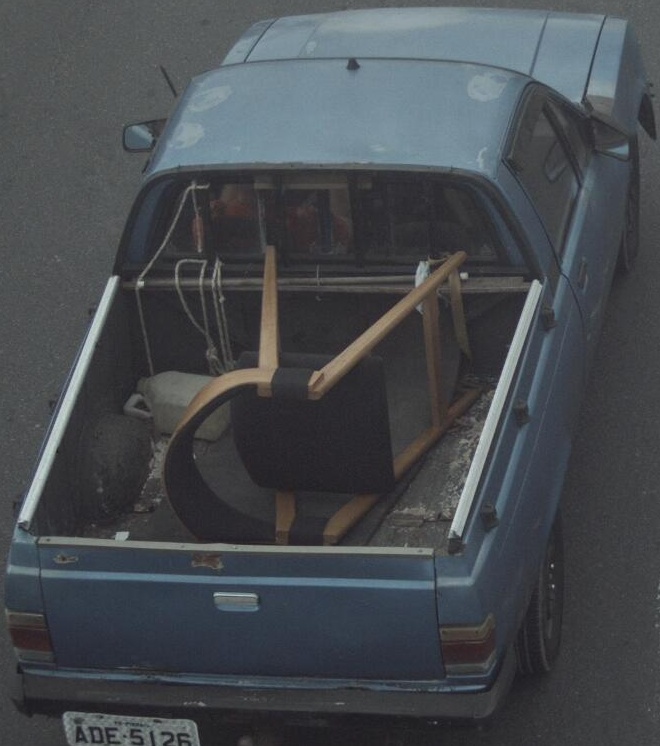}
    }%
    
    \vspace{1mm}
    
    \resizebox{0.925\linewidth}{!}{
        \includegraphics[height=8.5ex]{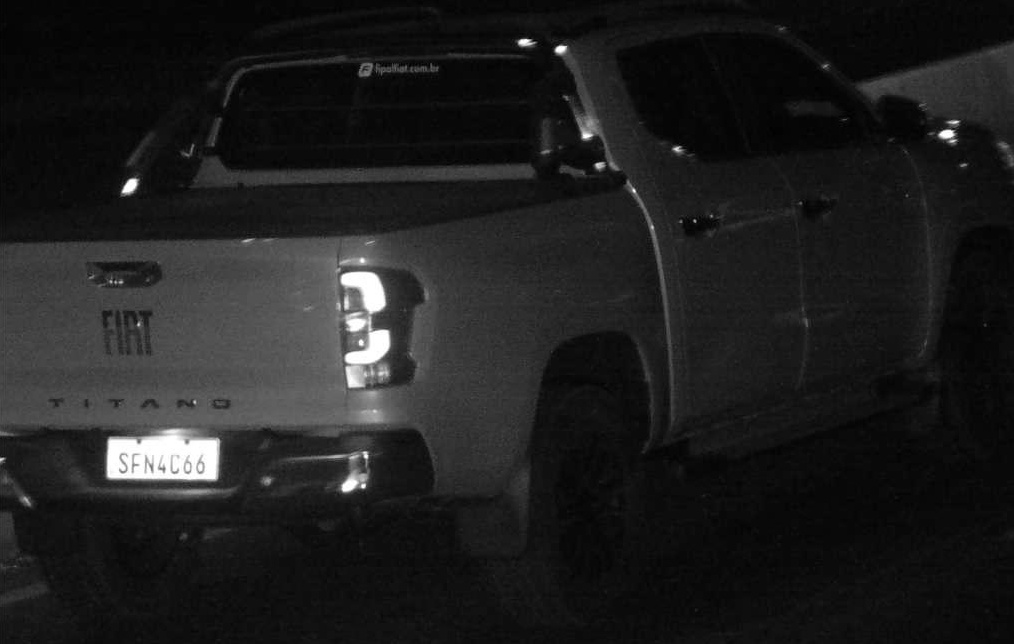}
        \includegraphics[height=8.5ex]{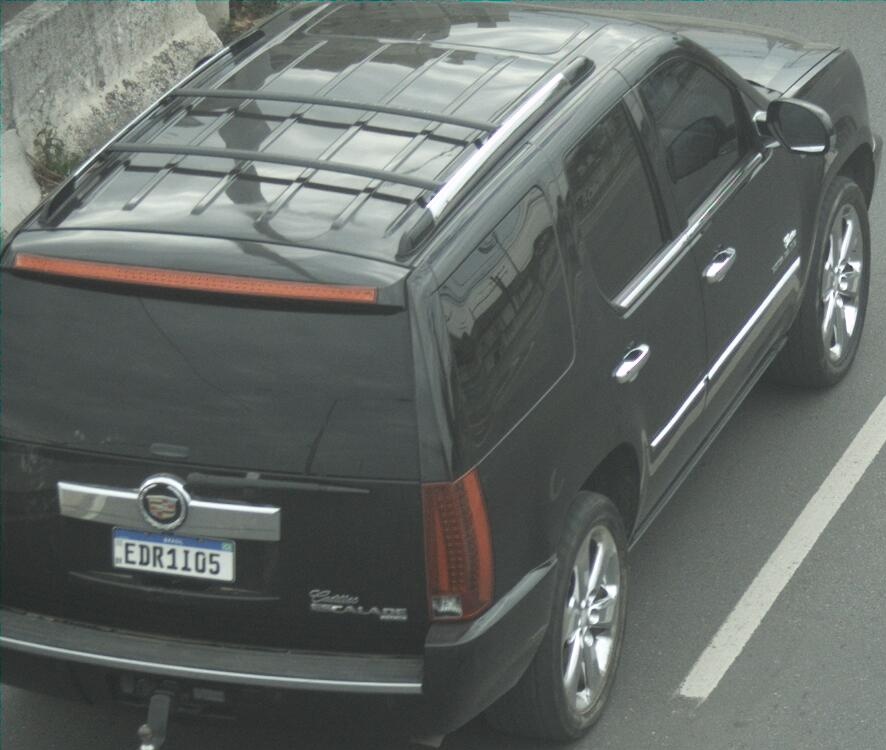}
        \includegraphics[height=8.5ex]{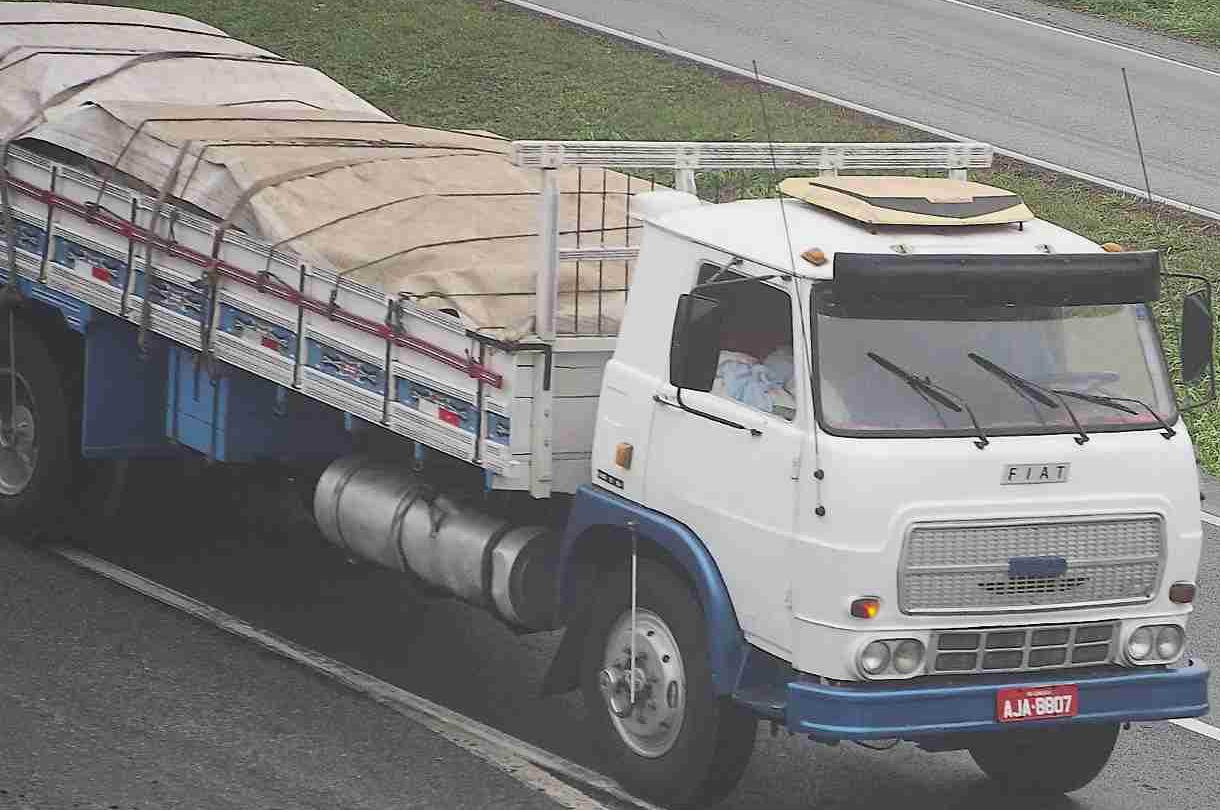}
        \includegraphics[height=8.5ex]{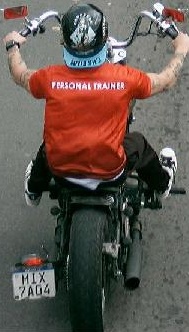}
    }%
    
    \vspace{-2mm}
    \caption{\major{Samples from the curated \acrfull{ood} set. These images introduce a distinct domain shift from the UFPR-VeSV dataset by presenting unseen vehicle models that map directly to its original type classes.}}  
    \label{fig:ood-set}
\end{figure}

\section{Results and Discussion}
\label{sec:results}

\major{\cref{tab:linear_probes} summarizes the linear probing performance across vehicle type, make, and model recognition tasks, comparing foundation models against EfficientNet-V2 --- the best-performing model from the original UFPR-VeSV work~\cite{lima2026toward}.}
Most notably, DINOv3 outperformed both the dataset baseline and all evaluated vision foundation models, achieving a \gls*{ma-acc} of $97.7$\% for vehicle type and over $93.0$\% for fine-grained make and model recognition. 
\major{These results demonstrate that DINOv3 provides highly discriminative frozen representations for all three recognition tasks. While DINOv3 is a massively scaled 2D self-supervised model, we do not attribute this advantage solely to its 2D training objective, as differences in pre-training data, model capacity, and architecture are also likely contributing factors.}
\major{A paired Wilcoxon signed-rank test across the 10 official splits confirmed that DINOv3 consistently outperformed the best-performing 3D-aware model, Depth Anything v2 ($p < 0.002$).}

\begin{table}[!htb]
    \centering
    \caption{Linear probe results (\%) for vehicle type, make, and model recognition on UFPR-VeSV. Results are averaged over $10$ runs using the official dataset splits, with standard deviations in~parentheses.}
    \label{tab:linear_probes}

    \vspace{-2mm}
    
    \small
    \setlength{\tabcolsep}{4pt}
    \resizebox{0.99\linewidth}{!}{
        \begin{tabular}{@{} l c c c c c c @{}}
            \toprule
            & \multicolumn{2}{c}{Type} & \multicolumn{2}{c}{Make} & \multicolumn{2}{c}{Model} \\
            \cmidrule(lr){2-3} \cmidrule(lr){4-5} \cmidrule(lr){6-7}
            Method & Ma-Acc & Mi-Acc & Ma-Acc & Mi-Acc & Ma-Acc & Mi-Acc \\
            \midrule
            \multicolumn{7}{@{}l}{\textit{\major{Best-performing model reported in the original work introducing UFPR-VeSV~\cite{lima2026toward}}}} \\
            \addlinespace
            EfficientNet-V2~\cite{lima2026toward} & $89.0$ $(2.0)$ & $96.1$ $(0.7)$ & $85.0$ $(1.6)$ & $94.4$ $(0.6)$ & $86.2$ $(1.1)$ & $90.9$ $(0.6)$ \\
            \midrule
            \multicolumn{7}{@{}l}{\textit{Standard 2D Vision Foundation Models}} \\
            \addlinespace
            DINOv1 & $91.0$ $(2.3)$ & $92.5$ $(0.7)$ & $67.4$ $(2.6)$ & $74.9$ $(2.4)$ & $66.9$ $(2.5)$ & $75.6$ $(1.9)$ \\
            iBOT (ImageNet-22K) & $90.4$ $(1.5)$ & $92.0$ $(0.9)$ & $63.9$ $(1.7)$ & $72.0$ $(2.9)$ & $63.9$ $(2.0)$ & $74.1$ $(1.5)$ \\
            DINOv2 & $95.3$ $(1.3)$ & $97.1$ $(0.4)$ & $65.9$ $(3.8)$ & $73.2$ $(1.6)$ & $66.5$ $(1.8)$ & $76.8$ $(1.3)$ \\
            DINOv3 & $\mathbf{97.7}$ $(1.0)$ & $\mathbf{98.7}$ $(0.2)$ & $\mathbf{93.5}$ $(1.1)$ & $\mathbf{97.5}$ $(0.2)$ & $\mathbf{95.0}$ $(0.7)$ & $\mathbf{94.7}$ $(0.6)$ \\
            SAM 3 & $83.9$ $(1.5)$ & $89.0$ $(0.6)$ & $32.6$ $(4.8)$ & $37.2$ $(2.6)$ & $27.6$ $(2.2)$ & $33.2$ $(2.7)$ \\
            \midrule
            \multicolumn{7}{@{}l}{\textit{3D-Aware Vision Foundation Models}} \\
            \addlinespace
            CroCo v2 & $72.0$ $(2.9)$ & $75.2$ $(1.0)$ & $26.9$ $(5.0)$ & $28.5$ \major{$(2.2)$} & $20.1$ $(0.8)$ & $21.1$ $(1.9)$ \\
            Depth Anything v1 & $94.5$ $(1.4)$ & $96.2$ $(0.4)$ & $\mathbf{58.2}$ $(3.5)$ & $\mathbf{62.2}$ $(1.9)$ & $61.1$ $(1.2)$ & $\mathbf{68.8}$ $(1.8)$ \\
            Depth Anything v2 & $\mathbf{95.0}$ $(1.7)$ & $\mathbf{96.4}$ $(0.4)$ & $56.6$ $(4.7)$ & $59.2$ $(2.6)$ & $\mathbf{64.4}$ $(1.6)$ & $68.1$ $(2.0)$ \\
            DUSt3R (Encoder) & $74.7$ $(1.9)$ & $79.0$ $(1.6)$ & $29.1$ $(3.7)$ & $31.0$ $(1.6)$ & $25.0$ $(1.4)$ & $26.4$ $(2.1)$ \\
            Gamba & $37.3$ $(1.7)$ & $61.7$ $(3.0)$ & $11.0$ $(2.3)$ & $22.2$ $(1.6)$ & $\phantom{0}5.1$ $(1.0)$ & $\phantom{0}9.0$ $(1.3)$ \\
            Hunyuan3D 2.1 & $62.6$ $(4.3)$ & $77.0$ $(1.4)$ & $15.8$ $(1.8)$ & $18.8$ $(1.9)$ & $12.6$ $(1.4)$ & $16.4$ $(0.8)$ \\
            MASt3R (Encoder) & $72.0$ $(3.2)$ & $76.6$ $(0.8)$ & $24.5$ $(4.7)$ & $29.0$ $(1.6)$ & $19.3$ $(0.8)$ & $20.8$ $(1.9)$ \\
            SAM 3D & $17.3$ $(0.8)$ & $62.2$ $(1.6)$ & $\phantom{0}7.1$ $(0.2)$ & $22.5$ $(1.6)$ & $\phantom{0}3.0$ $(0.4)$ & $\phantom{0}2.2$ $(0.5)$ \\
            Depth Anything 3 & $24.1$ $(3.1)$ & $40.3$ $(5.7)$ & $\phantom{0}7.4$ $(1.9)$ & $\phantom{0}9.2$ $(4.6)$ & $\phantom{0}2.1$ $(0.7)$ & $\phantom{0}1.9$ $(0.7)$ \\
            \bottomrule
        \end{tabular}
    }
\end{table}

\major{The evaluated 3D-aware models showed a distinct disparity depending on task granularity.}
For coarse vehicle type classification, models that use semantic priors with depth estimation (e.g., Depth Anything v1 and v2) yielded competitive results ($94.5$\% and $95.0$\% Ma-Acc, respectively). 
This aligns with the intuition that a vehicle's broad category correlates with its global 3D bounding volume. 
However, this geometric advantage collapses on fine-grained tasks. 

\blue{Architectures that abstract visual data through 3D reconstruction (e.g., SAM~3D, Gamba, DUSt3R) or image segmentation (e.g., SAM~3) struggled significantly with fine-grained recognition. 
This drop in performance suggests two possibilities: either these models suffer from a representational bottleneck, discarding high-frequency details to achieve spatial compression, or their reconstruction objectives produce highly entangled feature spaces.}

\major{As vehicle type recognition demonstrated the most promising performance, our error analysis focuses on this task for the top-performing 2D (DINOv3) and 3D-aware (Depth Anything v2) architectures. DINOv3 struggled with the \textit{tractor-truck} class, misclassifying $12\%$ of samples as standard \textit{trucks}. 
Although visually similar, these classes differ in size and geometry, explaining why the 3D-aware model avoided this error. 
Instead, Depth Anything v2 struggled with the \textit{minibus} class, misclassifying $33\%$ as \textit{buses}; their shared box-like geometry likely hinders the model's geometric priors.}

\major{We also analyzed the impact of vehicle viewpoint on classification accuracy. 
A paired Wilcoxon signed-rank test reveals that the impact of pose differs significantly between the two models. 
While DINOv3 is vulnerable to changes in viewing angle ($p < 0.01$), Depth Anything v2 demonstrates stronger invariance to pose, showing no significant effect on Type classification ($p \approx 0.19$).}

\begin{table}[htb]
    \centering
    \caption{Non-linear probe results~(\%) on UFPR-VeSV. Results show the average and standard deviation (in parentheses) across 10 splits for the best 2D and 3D-aware vision foundation models.}
    \label{tab:non-linear}

    \vspace{-2mm}
    \setlength{\tabcolsep}{4pt}
    \resizebox{0.99\linewidth}{!}{
        \begin{tabular}{@{} l cc cc cc @{}}
            \toprule
            & \multicolumn{2}{c}{Type} & \multicolumn{2}{c}{Make} & \multicolumn{2}{c}{Model} \\
            \cmidrule(lr){2-3} \cmidrule(lr){4-5} \cmidrule(lr){6-7}
            Model & Ma-Acc & Mi-Acc & Ma-Acc & Mi-Acc & Ma-Acc & Mi-Acc \\
            \midrule
            \multicolumn{7}{@{}l}{\textit{Best 3D Vision Foundation Model}} \\
            \addlinespace
            Depth Anything v2 &
            $\textbf{95.0}$ $\textbf{(1.7)}$ & $\textbf{97.1}$ $\textbf{(0.4)}$ &
            $67.8$ $(2.9)$ & $77.1$ $(0.3)$ &
            $72.4$ $(1.0)$ & $76.2$ $(1.6)$ \\
            \midrule
            \multicolumn{7}{@{}l}{\textit{Best 2D-Aware Vision Foundation Model}} \\
            \addlinespace
            DINOv3 &
            $94.8$ $(2.2)$ & $96.0$ $(0.4)$ &
            $\mathbf{92.2}$ $\mathbf{(1.6)}$ & $\mathbf{96.0}$ $\mathbf{(0.4)}$ &
            $\mathbf{93.9}$ $\mathbf{(1.0)}$ & $\mathbf{94.0}$ $\mathbf{(0.7)}$ \\
            \bottomrule
        \end{tabular}
    }
\end{table}

\major{Concluding the benchmark analysis, \cref{tab:non-linear} presents the non-linear probing results for DINOv3 and Depth Anything v2, the best-performing architectures. 
The results shows that while the non-linear setup improves Depth Anything v2 and slightly degrades DINOv3, DINOv3 still heavily outperforms Depth Anything v2 on Make and Model recognition. 
This suggests that DINOv3 produces robust and linearly separable features that do not require non-linear classifiers to achieve top performance in the studied~scenario.}

\begin{figure}[t]
    \centering
    
    \resizebox{0.9\linewidth}{!}{
        \includegraphics{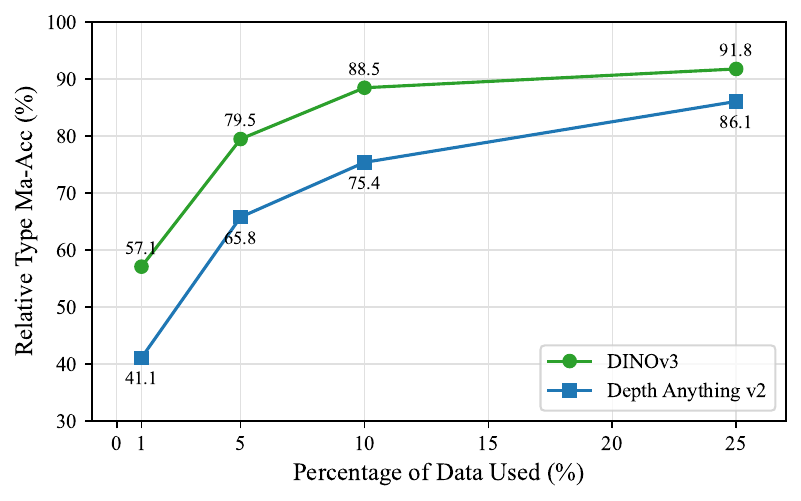}
    }
    \vspace{-1mm}
    
    \caption{Few-shot efficiency evaluation of DINOv3 and Depth Anything v2, trained using linear-probing on different proportions of the UFPR-VeSV dataset.}
    \label{fig:sample-efficiency-graph}
\end{figure}

\major{Next, we evaluate data efficiency. \cref{fig:sample-efficiency-graph} illustrates the classification accuracy of DINOv3 and Depth Anything v2 with linear probing trained on $1\%$, $5\%$, $10\%$, and $25\%$ data subsets. 
Although performance drops across when compared to the full-dataset baseline, DINOv3 maintains a $91.8\%$ \gls*{ma-acc} for vehicle type recognition with only $25\%$ of the data. 
This constrained performance still rivals the full-dataset results of many 3D-aware architectures, underscoring the efficiency of scaled 2D representations. 
Meanwhile, Depth Anything v2 degrades more sharply under extreme scarcity.}

\major{Continuing the robustness analysis, \cref{tab:ood_generalization} presents the \gls{ood} generalization results. 
A theoretical advantage of explicitly 3D-aware models is their ability to encode underlying geometry independently of superficial visual textures. 
Nonetheless, the performance trends observed in the linear probing benchmark persist. 
A paired Wilcoxon signed-rank test reveals that DINOv3 achieves higher Ma-Acc than Depth Anything v2 ($p < 0.01$), while there is no significant difference in their Mi-Acc ($p \approx 0.084$). Furthermore, comparing the original DINOv3 linear probe with the Depth Anything v2 MLP reveals that the DINOv3 linear probe continues to outperform Depth Anything v2.} 

\begin{table}[htb]
    \centering
    \caption{Out-of-Distribution (OOD) classification accuracy for vehicle type recognition on unseen vehicle data.}
    \label{tab:ood_generalization}

    \vspace{-2mm}
    
    \begin{tabular}{@{} l c c @{}}
        \toprule
        Model & Ma-Acc (\%) & Mi-Acc  (\%) \\
        \midrule
        DINOv3 & $89.6$ $(0.8)$ & $89.7$ $(0.9)$  \\
        Depth Anything v2  & $84.6$ $(3.3)$ & $88.9$ $(1.1)$ \\
        \bottomrule
    \end{tabular}
\end{table}

\major{Concluding, standard 2D architectures achieved strong results across all tasks and experimental setups. 
However, it is important to contextualize this success. 
The top-performing model (i.e., DINOv3) benefits from pre-training on a massive data scale, which is likely a driving factor behind its robustness and linearly separable feature space. 
Therefore, these results should be interpreted with caution, as the compared models differ along dimensions beyond their 2D or 3D representational priors. 
Nonetheless, the evidence suggests that 3D-aware foundation models still need to be improved before they can be properly transferred beyond 3D downstream tasks.}

\section{Conclusions}
\label{sec:conclusion}

\glsresetall

This study presented an empirical benchmark evaluating the transferability of 2D and 3D-aware vision foundation models for vehicle attribute recognition. By assessing $14$ models as frozen feature extractors via linear probing on the challenging UFPR-VeSV~\cite{lima2026toward} dataset, we evaluated the inherent discriminative capabilities of their representations. Our results indicate that standard 2D self-supervised architectures currently provide the most effective features for these classification tasks, achieving the highest accuracy across vehicle type, make, and model recognition without requiring task-specific fine-tuning. 

Conversely, 3D-aware foundation models underperformed when used for vehicle recognition. We attribute this limitation to their pre-training objectives --- such as explicit 3D reconstruction and depth estimation --- which tend to abstract away the high-frequency details necessary for distinguishing specific vehicle identities. Even in \gls*{ood} scenarios, where geometric representations theoretically should offer better resilience to domain shifts, the \major{expected} advantage of 3D-aware models \major{did not materialize}.

Consequently, we conclude that current 3D-aware foundation models require dedicated adaptation strategies to become viable in this domain. 
\major{Crucially, however, this benchmark does not strictly isolate the effect of geometric priors from confounding variables, such as differences in pretraining data scale, parameter counts, and architectural design. 
Therefore, the observed performance gap should be interpreted as an empirical snapshot of current state-of-the-art models, rather than definitive evidence that 2D representations are inherently~superior.}

Future work should explore parameter-efficient fine-tuning methods to better adapt geometric priors to fine-grained tasks. \major{A more controlled comparison would require models sharing the same backbone, parameter count, and pre-training corpus while differing only in the incorporation of geometric supervision. Such controlled pre-training remains largely unavailable in the current foundation-model ecosystem and constitutes an important direction for future work.} Furthermore, systematically investigating the explicit impact of pre-training data scale across paradigms and developing hybrid architectures that leverage both 2D semantic richness and 3D spatial consistency is essential for advancing robust vehicle recognition systems.

\section*{\uppercase{Acknowledgments}}

\iffinal
    This study was financed in part by the \textit{Coordenação de Aperfeiçoamento de Pessoal de Nível Superior - Brasil~(CAPES)}, through the \textit{Programa de Excelência Acadêmica~(PROEX)} - Finance Code 001, in part by the \textit{Conselho Nacional de Desenvolvimento Científico e Tecnológico~(CNPq)} (\#~315409/2023-1), and in part by the \textit{Fundação Araucária}~(\#~078/2026).
\fi

\iffinal
    \balance
\else
\fi

\bibliographystyle{IEEEtran}
\bibliography{bibtex}

@article{amir2021deep,
  author    = {Shir Amir and Yossi Gandelsman and Shai Bagon and Tali Dekel},
  title     = {Deep {ViT} Features as Dense Visual Descriptors}, 
  journal   = {European Conference on Computer Vision (ECCV) --- What is Motion For? Workshop},
  year      = {2022},
}

@article{simeoni2025dinov3,
  title = {{DINOv3}},
  author = {Sim{\'e}oni, Oriane and others},
  journal = {arXiv preprint arXiv:2508.10104},
  year = {2025}
}

@article{oquab2023dinov2,
  title   = {{DINOv2}: Learning Robust Visual Features without Supervision},
  author  = {Oquab, Maxime and others},
  journal = {arXiv preprint arXiv:2304.07193},
  year    = {2023}
}

@inproceedings{mast3r_eccv24,
  title = {Grounding Image Matching in {3D} with {MASt3R}},
  author = {Leroy, Vincent and Cabon, Yohann and Revaud, Jerome},
  year = {2024},
  booktitle = {European Conference on Computer Vision (ECCV)},
  pages = {71-91},
  doi = {10.1007/978-3-031-73220-1_5},
  isbn = {978-3-031-73220-1}
}

@article{carion2025sam3segmentconcepts,
  title   = {{SAM 3}: Segment Anything with Concepts},
  author  = {Carion, Nicolas and others},
  journal = {arXiv preprint arXiv:2511.16719},
  year    = {2025},
  doi     = {10.48550/arXiv.2511.16719}
}

@inproceedings{depth_anything_v2,
 author = {Yang, Lihe and Kang, Bingyi and Huang, Zilong and Zhao, Zhen and Xu, Xiaogang and Feng, Jiashi and Zhao, Hengshuang},
 booktitle = {International Conference Neural Information Processing Systems (NeurIPS)}, 
 doi = {10.52202/079017-0688},
 pages = {21875-21911},
 publisher = {},
 title = {Depth Anything {V2}},
 volume = {37},
 year = {2024}
}

@article{depthanything3,
  title={Depth Anything 3: Recovering the visual space from any views},
  author={Haotong Lin and Sili Chen and Jun Hao Liew and Donny Y. Chen and Zhenyu Li and Guang Shi and Jiashi Feng and Bingyi Kang},
  journal={arXiv preprint arXiv:2511.10647},
  year={2025}
}

@inproceedings{grill2020byol,
author = {Grill, Jean-Bastien and others},
title = {Bootstrap your own latent a new approach to self-supervised learning},
year = {2020},
isbn = {9781713829546},
publisher = {},
address = {},
pages={1-14},
booktitle = {International Conference on Neural Information Processing Systems (NeurIPS)},
articleno = {1786},
numpages = {14},
location = {Vancouver, BC, Canada},
series = {}
}

@inproceedings{croco_v2,
  title = {{CroCo v2}: Improved Cross-view Completion Pre-training for Stereo Matching and Optical Flow},
  author = {Weinzaepfel, Philippe and others},
  year = {2023},
  booktitle = {IEEE/CVF International Conference on Computer Vision (ICCV)},
  volume = {},
  number = {},
  pages = {17923-17934},
  doi = {10.1109/ICCV51070.2023.01647}
}

@InProceedings{sam3dteam2025sam3d3dfyimages,
    author    = {Chen, Xingyu and others},
    title     = {{SAM 3D: 3Dfy} Anything in Images},
    booktitle = {IEEE/CVF Conference on Computer Vision and Pattern Recognition (CVPR)},
    year      = {2026},
    pages     = {7220-7232}
}

@article{shen2025gamba,
  title = {Gamba: Marry Gaussian Splatting With Mamba for Single-View {3D} Reconstruction},
  author = {Shen, Qiuhong and Wu, Zike and Yi, Xuanyu and Zhou, Pan and Zhang, Hanwang and Yan, Shuicheng and Wang, Xinchao},
  year = {2025},
  journal = {IEEE Transactions on Pattern Analysis and Machine Intelligence},
  volume = {},
  number = {},
  pages = {1-14},
  doi = {10.1109/TPAMI.2025.3569596}
}

@article{zhou2021ibot,
  title={{iBOT}: Image {BERT} Pre-Training with Online Tokenizer},
  author={Zhou, Jinghao and Wei, Chen and Wang, Huiyu and Shen, Wei and Xie, Cihang and Yuille, Alan and Kong, Tao},
  journal={International Conference on Learning Representations (ICLR)},
  year={2022},
  pages={1-29}
}

@inproceedings{depth_anything_v1,
  title = {Depth Anything: Unleashing the Power of Large-Scale Unlabeled Data},
  author = {Yang, Lihe and Kang, Bingyi and Huang, Zilong and Xu, Xiaogang and Feng, Jiashi and Zhao, Hengshuang},
  year = {2024},
  booktitle = {IEEE/CVF Conference on Computer Vision and Pattern Recognition (CVPR)},
  volume = {},
  number = {},
  pages = {10371--10381},
  doi = {10.1109/CVPR52733.2024.00987}
}

@inproceedings{dust3r_cvpr24,
  title = {{DUSt3R}: Geometric {3D} Vision Made Easy},
  author = {Wang, Shuzhe and Leroy, Vincent and Cabon, Yohann and Chidlovskii, Boris and Revaud, Jerome},
  year = {2024},
  booktitle = {IEEE/CVF Conference on Computer Vision and Pattern Recognition (CVPR)},
  volume = {},
  number = {},
  pages = {20697-20709},
  doi = {10.1109/CVPR52733.2024.01956}
}

@article{bommasani2022opportunitiesrisksfoundationmodels,
  title   = {On the Opportunities and Risks of Foundation Models},
  author  = {Bommasani, Rishi and others},
  journal = {arXiv preprint arXiv:2108.07258},
  year    = {2022},
  doi     = {10.48550/arXiv.2108.07258},
}

@inproceedings{caron2021dino,
  title = {Emerging Properties in Self-Supervised Vision Transformers},
  author = {Caron, Mathilde and Touvron, Hugo and Misra, Ishan and Jegou, Hervé and Mairal, Julien and Bojanowski, Piotr and Joulin, Armand},
  year = {2021},
  booktitle = {IEEE/CVF International Conference on Computer Vision (ICCV)},
  volume = {},
  number = {},
  pages = {9630-9640},
  doi = {10.1109/ICCV48922.2021.00951}
}

@article{hunyuan3d2025hunyuan3d,
  title   = {{Hunyuan3D 2.1}: From Images to High-Fidelity 3D Assets with Production-Ready {PBR} Material},
  author  = {{Tencent Hunyuan3D Team}},
  journal = {arXiv preprint arXiv:2506.15442},
  year    = {2025},
  doi     = {10.48550/arXiv.2506.15442}
}

@article{zhang2013reliable,
    author={Zhang, Bailing},
    journal={IEEE Transactions on Intelligent Transportation Systems}, 
    title={Reliable Classification of Vehicle Types Based on Cascade Classifier Ensembles}, 
    volume={14},
    number={1},
    pages={322-332},
    year={2013},
    doi={10.1109/TITS.2012.2213814}
}

@article{dong2015vehicle,
    author={Dong, Zhen and Wu, Yuwei and Pei, Mingtao and Jia, Yunde},
    journal={IEEE Transactions on Intelligent Transportation Systems}, 
    title={Vehicle Type Classification Using a Semisupervised Convolutional Neural Network}, 
    year={2015},
    volume={16},
    number={4},
    pages={2247-2256},
    doi={10.1109/TITS.2015.2402438}
}

@article{sochor2019boxcars,
    author={Sochor, Jakub and Špaňhel, Jakub and Herout, Adam},
    journal={IEEE Transactions on Intelligent Transportation Systems}, 
    title={{BoxCars}: Improving Fine-Grained Recognition of Vehicles Using 3-{D} Bounding Boxes in Traffic Surveillance}, 
    volume={20},
    number={1},
    pages={97-108},
    year={2019},
    doi={10.1109/TITS.2018.2799228}
}

@article{he2024vehicle,
  author={He, Chunguang and Wang, Dianhai and Cai, Zhengyi and Zeng, Jiaqi and Fu, Fengjie},
  journal={IEEE Transactions on Intelligent Transportation Systems}, 
  title={A Vehicle Matching Algorithm by Maximizing Travel Time Probability Based on Automatic License Plate Recognition Data}, 
  year={2024},
  volume={25},
  number={8},
  pages={9103-9114},
  doi={10.1109/TITS.2024.3358625}
}

@article{gayen2025simsanet,
    author = {Gayen, Soumyajit and Maity, Sourajit and Singh, Pawan Kumar and Sarkar, Ram},
    title = {{SimSANet}: a simple sequential attention-aided deep neural network for vehicle make and model recognition},
    journal = {Neural Computing and Applications},
    volume = {37},
    number = {1},
    pages = {319–339},
    year = {2025},
    doi = {10.1007/s00521-024-10480-z}
}

@article{sathyam2025foundation,
    author={Sathyam, Rajendramayavan and Li, Yueqi},
    journal={IEEE Open Journal of Vehicular Technology}, 
    title={Foundation Models for Autonomous Driving Perception: A Survey Through Core Capabilities}, 
    year={2025},
    volume={6},
    number={},
    pages={2554-2582},
    doi={10.1109/OJVT.2025.3604823}
}

@article{laroca2025advancing,
    title = {Advancing Multinational License Plate Recognition Through Synthetic and Real Data Fusion: A Comprehensive Evaluation},
    author = {{Laroca}, Rayson and {Estevam}, Valter and {Moreira}, Gladston J. P. and {Minetto}, Rodrigo and {Menotti}, David},
    year = {2025},
    journal = {IET Intelligent Transport Systems},
    volume = {19},
    number = {1},
    pages = {e70086},
    doi = {10.1049/itr2.70086}
}

@article{li2025weakly,
    author={Li, Linhao and Zang, Han and Fan, Xiaojuan and Cheng, Hao and Dong, Yongfeng},
    journal={IEEE Transactions on Intelligent Transportation Systems}, 
    title={Weakly Supervised Bilinear Convolutional Neural Network for Fine-Grained Vehicle Classification}, 
    volume={26},
    number={12},
    pages={22470-22481},
    year={2025},
    doi={10.1109/TITS.2025.3621831}
}

@article{lu2025automated,
    author = {Linjun Lu  and Fei Dai },
    title = {Automated {FHWA} Vehicle Classification Using Combined Semantic and Geometric Features Extracted from Surveillance Videos},
    journal = {Journal of Computing in Civil Engineering},
    volume = {39},
    number = {3},
    pages = {04025023},
    year = {2025},
    doi = {10.1061/JCCEE5.CPENG-6413}
}

@article{munoz2025veri,
  title={{Veri-Car}: Towards open-world vehicle information retrieval},
  author={Munoz, Andr{\'e}s and Thomas, Nancy and Vapsi, Annita and Borrajo, Daniel},
  journal={Neural Computing and Applications},
  volume={37},
  number={20},
  pages={15183--15221},
  year={2025},
  publisher={Springer}
}

@article{nascimento2025toward,
  title = {Toward Advancing License Plate Super-Resolution in Real-World Scenarios: A Dataset and Benchmark},
  author = {V. {Nascimento} and G. E. {Lima} and R. O. {Ribeiro} and W. R. {Schwartz} and R. {Laroca} and D. {Menotti}},
  year = {2025},
  journal = {Journal of the Brazilian Computer Society},
  volume = {1},
  number = {31},
  pages = {435-449},
  doi = {10.5753/jbcs.2025.5159},
  issn = {},
}

@article{tan2025cross,
    author={Tan, Shi Hao and Chuah, Joon Huang and Chow, Chee-Onn and Kanesan, Jeevan},
    journal={IEEE Transactions on Intelligent Transportation Systems}, 
    title={Cross-Granularity Network for Vehicle Make and Model Recognition}, 
    year={2025},
    volume={26},
    number={},
    pages={5782-5791},
    doi={10.1109/TITS.2025.3549217}
}

@article{wojcik2025lplc,
    title = {{LPLC}: A Dataset for License Plate Legibility Classification},
    author = {L. {Wojcik} and G. E. {Lima} and V. {Nascimento} and E. {Nascimento Jr.} and R. {Laroca} and D. {Menotti}},
    year = {2025},
    journal = {Conference on Graphics, Patterns and Images (SIBGRAPI)},
    volume = {},
    number = {},
    pages = {},
    doi = {10.1109/SIBGRAPI67909.2025.11223367},
    issn = {1530-1834}
}

@article{lima2026toward,
    author = {Gabriel Eduardo Lima and Valfride Nascimento and Eduardo Santos and Eduil Nascimento Jr. and Rayson Laroca and David Menotti},
    title = {Toward Unified Fine-Grained Vehicle Classification and Automatic License Plate Recognition},
    journal = {Journal of the Brazilian Computer Society},
    volume = {32},
    number = {1},
    year = {2026},
    pages = {783--799},
    doi = {10.5753/jbcs.2026.5899}
}

@article{wu2026perception,
    author={Wu, Wentao and Wang, Xiao and Li, Chenglong and Tang, Jin and Luo, Bin},
    journal={IEEE Transactions on Circuits and Systems for Video Technology}, 
    title={Vehicle-Centric Perception via Multimodal Structured Pre-Training}, 
    year={2026},
    volume={36},
    number={6},
    pages={8615-8630},
    doi={10.1109/TCSVT.2026.3663409}
}

@article{GAYEN2024101885,
title = {Two decades of vehicle make and model recognition – Survey, challenges and future directions},
journal = {Journal of King Saud University - Computer and Information Sciences},
volume = {36},
number = {1},
pages = {101885},
year = {2024},
issn = {1319-1578},
doi = {https://doi.org/10.1016/j.jksuci.2023.101885},
author = {Soumyajit Gayen and Sourajit Maity and Pawan Kumar Singh and Zong Woo Geem and Ram Sarkar}
}

@inproceedings{ma2005edge,
  author={Xiaoxu Ma and Grimson, W.E.L.},
  booktitle={IEEE International Conference on Computer Vision (ICCV)}, 
  title={Edge-based rich representation for vehicle classification}, 
  year={2005},
  volume={},
  number={},
  pages={1185-1192},
  doi={10.1109/ICCV.2005.80}
}

@inproceedings{chen2011vehicle,
    author={Chen, Zezhi and Ellis, Tim and Velastin, Sergio A},
    booktitle={IEEE Conference on Intelligent Transportation Systems (ITSC)}, 
    title={Vehicle type categorization: A comparison of classification schemes}, 
    year={2011},
    pages={74-79},
    doi={10.1109/ITSC.2011.6083075}
}

@inproceedings{krause2013collecting,
    author    = {Krause, Jonathan and Deng, Jia and Stark, Michael and Fei-Fei, Li},
    title     = {Collecting a Large-Scale Dataset of Fine-Grained Cars},
    booktitle = {IEEE Conference on Computer Vision and Pattern Recognition (CVPR) --- FGVC Workshop},
    year      = {2013},
}

@inproceedings{krause2013object,
    author={Krause, Jonathan and Stark, Michael and Deng, Jia and Fei-Fei, Li},
    booktitle={IEEE International Conference on Computer Vision Workshops}, 
    title={3{D} Object Representations for Fine-Grained Categorization}, 
    year={2013},
    volume={},
    number={},
    pages={554-561},
    doi={10.1109/ICCVW.2013.77}
}

@inproceedings{lin2014jointly,
    title={Jointly optimizing {3D} model fitting and fine-grained classification},
    author={Lin, Yen-Liang and Morariu, Vlad I and Hsu, Winston and Davis, Larry S},
    booktitle={European Conference on Computer Vision (ECCV)},
    pages={466-480},
    year={2014},
    doi={10.1007/978-3-319-10593-2_31}
}

@inproceedings{yang2015compcars,
    author={Yang, Linjie and Luo, Ping and Tang, Xiaoou},
    booktitle={IEEE Conference on Computer Vision and Pattern Recognition (CVPR)}, 
    title={A large-scale car dataset for fine-grained categorization and verification}, 
    year={2015},
    volume={},
    number={},
    pages={3973-3981},
    doi={10.1109/CVPR.2015.7299023}
}

@misc{senatran_frota_2026,
  author       = {{Secretaria Nacional de Trânsito (SENATRAN)}},
  title        = {Frota de Veículos 2026},
  year         = {2026},
  howpublished = {\url{https://www.gov.br/transportes/pt-br/assuntos/transito/conteudo-Senatran/frota-de-veiculos-2026}},
  note         = {Accessed: 2026-05-07}
}

@inproceedings{laroca2026competition,
  title = {{ICPR} 2026 {C}ompetition on Low-Resolution License Plate Recognition},
  author = {R. {Laroca} and others},
  year = {2026},
  month = {Aug},
  booktitle = {International Conference on Pattern Recognition (ICPR)},
  pages = {256-275},
  doi = {10.1007/978-3-032-31936-4\_18},
  isbn = {978-3-032-31936-4}
}

@inproceedings{orru2026revisiting,
  title = {Revisiting Vehicle Color Recognition in Long-Tailed Surveillance Scenarios},
  author = {V. {Orrú} and B. H. {Foggiatto} and G. E. {Lima} and D. {Menotti} and R. {Laroca}},
  year = {2026},
  month = {Aug},
  booktitle = {International Conference on Pattern Recognition (ICPR) -- V3SC Workshop},
  pages = {1-15},
  doi = {},
  issn = {}
}

@inproceedings{yao2025labelany3d,
    title={{LabelAny3D}: Label Any Object 3D in the Wild},
    author={Jin Yao and Radowan Mahmud Redoy and Sebastian Elbaum and Matthew B. Dwyer and Zezhou Cheng},
    booktitle={International Conference on Neural Information Processing Systems (NeurIPS)},
    pages={133188--133216},
    year={2025}
}

\end{document}